\documentclass[final]{format/cvpr}

\usepackage{times}
\usepackage{epsfig}
\usepackage{graphicx}
\usepackage{amsmath}
\usepackage{amssymb}
\usepackage{comment}
\usepackage{float}

\usepackage[USenglish]{babel}
\usepackage[nolist]{acronym}
\usepackage{adjustbox} 
\usepackage{environ}   
\usepackage{booktabs}  
\usepackage{tabulary}
\usepackage{multirow}
\usepackage{makecell}
\usepackage{balance}
\usepackage{cite} 

\usepackage{siunitx} 
\usepackage{pifont} 
\newcommand{\cmark}{\ding{51}} 
\newcommand{\xmark}{\ding{55}} 

\usepackage{soul} 
\DeclareRobustCommand{\hlLH}[1]{{\sethlcolor{cyan}\hl{#1}}}

\usepackage[pagebackref=true,breaklinks=true,colorlinks,bookmarks=false]{hyperref}

\usepackage[capitalize,
            nameinlink,
            noabbrev
            ]{cleveref}

\def\cvprPaperID{****} 
\def\confYear{CVPR 2021}

\begin{document}

\title{Measuring what Really Matters: Optimizing Neural Networks for TinyML}

\author{Lennart Heim\\
ETH Zürich\\
RWTH Aachen\\
{\tt\small leheim@ethz.ch}
\and
Andreas Biri\\
ETH Zürich\\
{\tt\small abiri@ethz.ch}
\and
Zhongnan Qu\\
ETH Zürich\\
{\tt\small quz@ethz.ch}
\and
Lothar Thiele\\
ETH Zürich\\
{\tt\small thiele@ethz.ch}
}

\maketitle


\begin{abstract}

    With the surge of inexpensive computational and memory resources, \acp{NN} have experienced an unprecedented growth in architectural and computational complexity. 
    Introducing \acp{NN} to resource-constrained devices 
    enables cost-efficient deployments, widespread availability, and the preservation of sensitive data.

    This work addresses the challenges of bringing Machine Learning to \acp{MCU}, where we focus on the ubiquitous ARM Cortex-M architecture.  
    The detailed effects and trade-offs that optimization methods, software frameworks, and \ac{MCU} hardware architecture have on key performance metrics such as inference latency and energy consumption have not been previously studied in depth for state-of-the-art frameworks such as TensorFlow Lite Micro.
    We find that empirical investigations which measure the perceptible metrics --\textit{performance as experienced by the user}-- are indispensable, as the impact of specialized instructions and layer types can be subtle.
    To this end, we propose an implementation-aware design as a cost-effective method for verification and benchmarking.
    Employing our developed toolchain, we demonstrate how existing \ac{NN} deployments on resource-constrained devices can be improved by systematically optimizing \acp{NN} to their targeted application scenario.

\end{abstract}


\section{Introduction}\label{sec:intro}

The popularity of the \ac{IoT} has resulted in a plethora of applications demanding intelligence ``at the edge'', ranging from human pose recognition~\cite{jiang3DHumanPose2020}
and crowd surveillance~\cite{yiEagleEyeWearableCamerabased2020} to natural hazard monitoring~\cite{meyerEventtriggeredNaturalHazard2019a}.
However, current state-of-the-art \acp{NN} often make significant demands on memory, computation, and energy \cite{strubell-etal-2019-energy,openaiAICompute2018,quDeepPartialUpdating2020}. This contradicts the resource-constrained nature of \ac{IoT} devices which only provide a low-power \ac{MCU}.    
In TinyML, researchers attempt to bring \acp{NN} to these edge devices by exploring more efficient models using \ac{NAS}~\cite{caiOnceforAllTrainOne2020,liberisConstrainedNeuralArchitecture2020} and compressing pre-trained \acp{NN} for efficient on-device inference through pruning~\cite{hanDeepCompressionCompressing2016,heChannelPruningAccelerating2017a}, quantization~\cite{quAdaptiveLossAwareQuantization2020,jacobQuantizationTrainingNeural2017,linFixedPointQuantization2016a}, or distillation~\cite{hintonDistillingKnowledgeNeural2015}.

For efficient inference, research has focused on minimizing metrics such as the memory footprint of the network as well as the number of \ac{MACC} operations and \acp{FLOP} while maintaining an acceptable accuracy.
For this optimization, the latter serve as proxies for the computational complexity, as they can be easily determined analytically based on the network architecture, even before the computationally intensive training~\cite{liberisConstrainedNeuralArchitecture2020,banburyMicroNetsNeuralNetwork2021}.
Especially for \acp{MCU} and other CPU architectures, it has been claimed that \ac{MACC} operations provide a better proxy than for co-processors like GPUs due to their simpler hardware architecture~\cite{liberisConstrainedNeuralArchitecture2020,banburyMicroNetsNeuralNetwork2021}.
Nonetheless, those metrics are only \textit{proxies} for the metrics the user is exposed to: the inference latency and energy consumption -- what we call \textit{perceptible metrics}.
While commonly used for the final assessment of a network's efficiency, our experimental results 
show that these proxies do not always correctly reflect the final runtime metrics, resulting in misleading conclusions and suboptimal automated search strategies.
However, most current designs still substitute selection requirements with these proxy metrics and do not take perceptible metrics into consideration. 
In addition, they do not exploit architectural \ac{MCU} features that heavily influence the reliability of such metrics, thereby missing the opportunity for a symbiotic hardware-aware \ac{NN} design.

In this work, we analytically and experimentally investigate the implications of perceptible metrics for \ac{NN} design.
We examine their correlation to other common metrics and propose concrete design guidelines for future networks targeted at edge devices. 
To the best of our knowledge, we are the first to take such empirical metrics directly into account and facilitate the targeted selection of suitable \acp{NN} for application-specific scenarios.

While the hardware architecture of \acp{MCU} is less complex than desktop-class CPUs and co-processors such as GPUs or \acp{TPU}~\cite{jouppiIndatacenterPerformanceAnalysis2017}
, their computational and memory resources are inherently limited.
Therefore, achieving an efficient \ac{NN} execution by leveraging architectural features is paramount. Given the simplified underlying architecture (flat memory hierarchies, few to no caches, and shallow pipelines), we can analytically understand its implications and exploit it more effectively. 
Focusing current efforts of \ac{ML} on the edge solely on co-processors would render many of the existing systems and even more upcoming ones unsuitable.
The deployment on already available \ac{MCU} architectures
can enable on-device inference \textit{today} on billions of commercial devices.
We demonstrate this by deriving design guidelines based on an analysis of the ubiquitous ARM Cortex-M~\cite{armltdCortexM} architecture, used in $4.4$ billion \acp{MCU} sold in the last quarter of 2020 alone~\cite{armltdNews}.

In particular, we make the following key contributions:
\begin{enumerate}
    \item We present a complete hardware and software toolchain that enables the implementation-aware investigation of key performance metrics such as inference latency and energy consumption. It includes portable benchmarking tools 
    to quantify, analyze and optimize \acp{NN} at layer granularity by deploying them directly on \acp{MCU} -- shown in \cref{sec:method}.
    \item We demonstrate that experimental investigations are indispensable, as estimating energy efficiency and inference latency analytically is prone to anomalies. The combination of optimizations leads to a non-uniform acceleration across the \ac{NN}, whose layers display an intricate interplay that is difficult to quantify without empirical validation -- which we focus on in \cref{sec:experiments}. 
    \item We show that the use of proxy metrics such as operations can be misleading and neglects actual hardware utilization. Leveraging architectural insights and using implementation-aware metrics
    while designing \acp{NN} permits us to fine-tune networks. This allows us to increase their computational complexity and hence potential for higher accuracy~\cite{openaiAICompute2018} while simultaneously reducing latency and energy consumption by following simple guidelines, as presented in \cref{sec:discussion}.
\end{enumerate}


\section{Related work}\label{sec:related}

Various research efforts have targeted \acp{NN} on resource-constrained devices.
We group existing work into network compression, efficient architecture design, and \ac{NAS}.

\textbf{Compression}
Quantization reduces the bit-width of \ac{NN} parameters, which permits a drastic reduction of the memory footprint~\cite{quAdaptiveLossAwareQuantization2020,jacobQuantizationTrainingNeural2017,linFixedPointQuantization2016a}. 
It has become a standard compression technique in TinyML due to its significant memory savings while usually having a negligible effect on accuracy~\cite{davidTensorFlowLiteMicro2021}.
Whereas quantization can in principle be used with any bit-width, e.g. 4~bit~\cite{bannerPostTraining4bit2019} or an adaptive bit-width~\cite{quAdaptiveLossAwareQuantization2020}, we focus on 8~bit quantization which is supported by most \acp{MCU}.
Unsupported bit-widths 
need to be emulated, resulting in inefficient hardware utilization~\cite{armltdCortexM, banburyMicroNetsNeuralNetwork2021}.

While weight sharing \cite{hanDeepCompressionCompressing2016,chenFittingSearchSpace2020} and network pruning \cite{heChannelPruningAccelerating2017a,zhuangDiscriminationawareChannelPruning2019,hanDeepCompressionCompressing2016}
have shown promising results for efficient inference, their usage on edge devices remains an open challenge as they are not yet supported by open-source frameworks.

\textbf{Efficient architectures}
An efficient base architecture for further pre-deployment optimizations can either be designed manually or systematically searched for.
With the rise of \acp{NN} on phones, novel architectures such as Mobile\-Nets~\cite{howardMobileNetsEfficientConvolutional2017}, SqueezeNet~\cite{iandolaSqueezeNetAlexNetlevelAccuracy2016},  SparseNet~\cite{liuSparseNetSparseDenseNet2018}, Shuffle\-Net~\cite{zhangShuffleNetExtremelyEfficient2017}, and EfficientNet~\cite{tanEfficientNetRethinkingModel2019,liSearchingFastModel2021} have been proposed.
Those \acp{NN} highlight a new trend aside from primarily focusing on the accuracy, as they also consider network complexity -- the interplay between accuracy, number of parameters, activations, and operations.
However, their applicability to TinyML is limited, as the resources on such systems are far more constrained than for mobile phones.

\textbf{Neural architecture search}
Instead of manually designing networks, \ac{NAS} finds such efficient architectures in an automated manner by systematically exploring the design space using a set of selected evaluation criteria.
A \ac{NAS} system can be split into three components: the search algorithm, search space, and evaluation strategy~\cite{elskenNeuralArchitectureSearch2019}. 
Whereas search algorithms can be partially reused from existing work
, the search space has to be redesigned and drastically reduced due to the limited memory. 
The evaluation strategy also requires adjustments to take the specific application requirements and concrete hardware constraints into account. 

Furthermore, as many TinyML systems are user-facing \ac{IoT} devices, it is oftentimes required to optimize for multiple metrics at the same time. Perceptible metrics, such as latency and energy consumption, are subject to specification requirements and often more expressive and constrained than accuracy.
Therefore, multi-objective optimization became more prominent and has been incorporated into the evaluation strategy~\cite{hsuMONASMultiObjectiveNeural2018,elskenNeuralArchitectureSearch2019}.
However, while acknowledging the importance of perceptible metrics, these \ac{NAS} designs use proxies like operations after having verified a linear relationship between them~\cite{banburyMicroNetsNeuralNetwork2021,liberisConstrainedNeuralArchitecture2020}.
In contrast, the use of latency lookup tables~\cite{wuFBNetHardwareAwareEfficient2019} based on previously measured executions directly integrates perceptible metrics.

Based on the evaluation strategy, we can divide \ac{NAS} approaches into four categories: no hardware influence~\cite{liuDARTSDifferentiableArchitecture2019}, the usage of proxies characterized in advance~\cite{caiOnceforAllTrainOne2020,liberisConstrainedNeuralArchitecture2020,linMCUNetTinyDeep2020a,banburyMicroNetsNeuralNetwork2021}, hardware-aware measurement and usage of perceptible metrics~\cite{wuFBNetHardwareAwareEfficient2019}, and hardware/software co-design~\cite{zhangWhenNeuralArchitecture2019}.
While hardware-aware \ac{NAS} has primarily targeted co-processors~\cite{jouppiIndatacenterPerformanceAnalysis2017,andriYodaNNUltraLowPower2016} (more than 75\,\% of published papers~\cite{benmezianeComprehensiveSurveyHardwareAware2021}), we argue that CPU architectures are a worthwhile target due to their ubiquitous availability in existing, deployed systems. 
Our work investigates the perceptible metrics of this device class and evaluates the applicability of proxies through experiments to showcase its true potential.


\begin{figure*}[t]
    \centering
    \includegraphics[width=\linewidth]{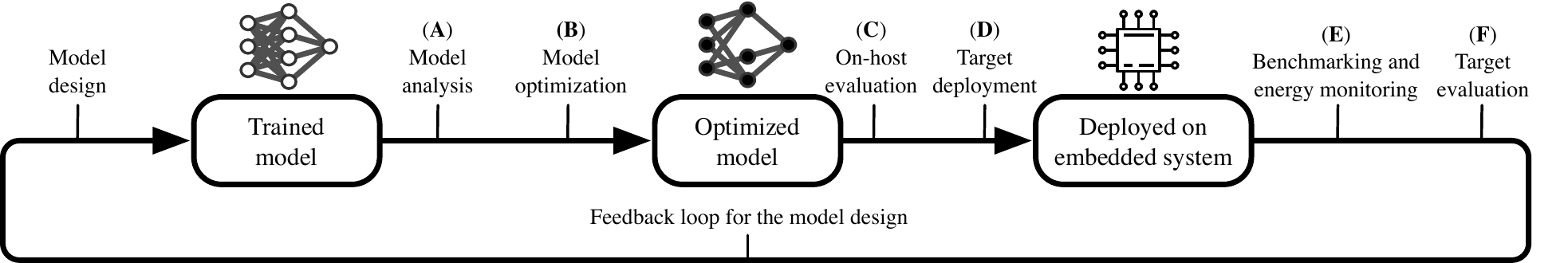}
    \caption{The toolchain includes portable benchmarking tools to quantify, analyze and optimize NNs by deploying them directly on MCUs. All stages can be used individually and allow an iterative design process.
    Our methodology enables the investigation of key performance metrics such as inference latency and energy consumption, gathering insights on the perceptible as well as standard metrics down to layer-wise granularity. An optional feedback of these results for the model design permits a cost-efficient, implementation-aware \ac{NAS}.}
    \label{fig:sec3-toolchain}
\end{figure*}

\section{Method}\label{sec:method}

In contrast to previous work which relied on proxy metrics, we focus on perceptible metrics which characterize the system as experienced by the user.
Therefore, they are the center of our designed methodology and resulting toolchain.

\subsection{Bridging theory and implementation}\label{sec:notallops}

By focusing on the number of operations as an optimization target, many details are lost in translation
~\cite{laiNotAllOps2018} and theoretical operation cost estimates do not necessarily match the atomic instructions of the hardware architecture which are eventually executed.
First, operating on floating point or on fixed point is a significant difference, especially when the underlying hardware does not natively support it (i.e. no FPU is available).
On top of this, the chosen bit-width is a key factor, as smaller bit-widths can be leveraged by larger bit-width architectures through aligned memory access and the usage of supported \ac{SIMD} instructions.
Aside from the quantization, the operation type is equally important. While a \ac{MACC} operation contains one multiplication and accumulation per cycle, \acp{FLOP} only execute a single computation. However, the exact conversion depends on the extent to which the underlying layer can exploit parallelization and is hence strongly implementation dependent.
Consequently, if 8~bit quantization is used on a 32~bit hardware architecture, an acceleration of up to $4\times$ is expected if aligned memory access and \ac{SIMD} can be leveraged.
However, the limited number of registers in \acp{MCU} and the NN structure can limit this exploitation, as we will discuss in \cref{sec:experiments}.

Taking these differences into account is intricate for complex architectures such as CPUs, GPUs, or TPUs~\cite{jouppiIndatacenterPerformanceAnalysis2017}.
Nonetheless, we find that the comparatively simple architecture of \acp{MCU}, in particular the shallow memory hierarchy and the predominantly serial computation, permits us to analyze and exploit it both analytically and experimentally.

To investigate the potential for exploitation, an empirical evaluation of the latency per \ac{MACC} operation enables an expressive comparison of different \acp{NN}, their layers, and target hardware.
In particular, the influence of optimizations on this perceptible metric demonstrates the variable comparative cost of individual layers.  
By calculating the number of operations for the layers of a \ac{NN} and measuring the inference latency, we can compute the resulting latency per operation for different layers and their hyper-parameters with
$ \delta = \frac{t_m}{c_e}$. $t_m$ represents the measured inference latency and $c_e$ is an estimate for the required MACC operations of the \ac{NN} layer (see \cref{sec:appendix-calcops} for the derivation).
As we disregard load and stores operations, this equation defines a upper bound 
on the actual execution latency.
However, we can extract knowledge on the relative efficiency of layer types themselves as well as the hyper-parameters' influence by comparing layers and different hyper-parameters~\cite{johnnyCMSISNNOptimizationsEdge2021}, which we will demonstrate in \cref{sec:experiments}.

\subsection{Implementation-aware toolchain}

Following established guidelines~\cite{banburyBenchmarkingTinyMLSystems2020}, our methodology allows us to gather first insights on the inference latency and energy consumption.
In addition to these perceptible metrics, we provide traditional metrics~\cite{szeHowEvaluateEfficient,szeEfficientProcessingDeep2020} to evaluate and design an architecture: accuracy, \ac{NN} size and computational complexity (i.e. number of operations). 
With this combination of metrics, we can observe the subtle impact of optimizations and investigate the correlation of standard metrics and their applicability for feedback on the architecture and optimizations under observation.
%
For this, we build on top of proven \ac{ML} methods and integrate insights from the embedded systems domain to design a toolchain that allows us to gather metrics down to layer-wise granularity.

In this subsection, we give a brief end-to-end overview of our proposed toolchain architecture, as depicted in \cref{fig:sec3-toolchain}. 
It consists of the the following components:
model analysis (\textbf{A}), optimizations (\textbf{B}), on-host evaluation (\textbf{C}), deployment (\textbf{D}), benchmarking and energy monitoring (\textbf{E}), and target evaluation (\textbf{F}).
Whereas components \textbf{A} - \textbf{C} are executed on the \textit{host} system (e.g., a desktop computer which is not resource-constrained), components \textbf{D} - \textbf{F} directly incorporate the targeted embedded system.
Notice that while we only depict the \textit{flow of information}, the design flow permits closer iterations as all of the components can be used individually to enable an efficient optimization process. 

\textbf{Model analysis (A)}
Starting with a previously designed and pre-trained base model that is not yet optimized for TinyML systems, the initial accuracy and loss can be easily determined.
Additionally, the memory footprint of the \ac{NN} (i.e. the required space for weights and biases) can be determined.
Those results can then be used as a reference for the impact of the optimizations on accuracy, loss, and model size.
Additionally, the model complexity can already be investigated at this stage, as the number of parameters as well as the estimated operations remain fixed.

\textbf{Optimizations (B)}
During the model optimization stage, we optimize the model using quantization as discussed in \cref{sec:related}.
At this stage, all optimizations are still hardware- and target-independent, as the deployment target is not yet known. However, the effect of quantization can depend on the deployment target and the supported bit-widths of the underlying MCU architecture. This can result in significant acceleration if the quantization type is chosen in synergy with the target hardware.

\textbf{On-host evaluation (C)}
Optimizations permanently alter the \ac{NN} and potentially influence its accuracy. Therefore, \textit{on-host} evaluation enables us to already assess the quantized model. The host system (e.g., a desktop computer) is not resource-constrained and permits a time-efficient evaluation using the test dataset -- which is usually too large to be stored directly on the resource-constrained system. 
There, the quantized model can be evaluated in regards to its accuracy and memory savings of the \ac{NN} parameters compared to the base model. As some models might suffer a more substantial accuracy loss than others through optimization, quantifying this dependence pre-deployment is crucial.

\textbf{Target deployment (D)}
Depending on the targeted device, specialized optimizations are available, such as software acceleration libraries like CMSIS-NN~\cite{laiCMSISNNEfficientNeural2018}, which provide the possibility to leverage dedicated, hardware-optimized kernels.
In contrast to the previously discussed optimizations, these do not change the numerical representation itself but increase the execution efficiency by leveraging target-specific features like SIMD instructions. 
During the deployment step, the already converted and optimized model and the required inference engine is compiled into the final firmware for the target processor.

\textbf{Benchmarking and energy monitoring (E)}
The benchmarking itself occurs on the target using the previously compiled benchmarking firmware and a dedicated energy measurement unit.
This benchmark enables us to gather the inference latency at layer resolution as well as the resulting classification of the on-device inferences.
We use a customized version of the firmware to measure the inference latency via software timers and leverage standard \ac{GPIO} signaling to interface with an external energy measurement unit.
As energy is a crucial contributor to deployment costs, its efficient usage plays a significant role in embedded systems.
To evaluate the energy consumption, a separate system that interfaces with and monitors the system under test is required to gather unbiased results based on known events, such as the start and stop of individual layers or the complete inference.

\textbf{Target evaluation (F)}
Lastly, to verify the previously gathered on-host classification accuracy results and the impact of target-specific hardware optimizations, we 
have the option to perform verification on the target itself.
As these device-dependent optimizations cannot be investigated before the deployment, the effective impact on the accuracy needs to be evaluated.
To do so, we send samples of the test dataset through a serial bus to the \ac{MCU}.
While the optimizations do not change the \ac{NN} itself, they might alter underlying computations depending on the specific implementation, potentially resulting in a different classification. Examples of this are the approximations of hyperbolic functions~\cite{laiCMSISNNEfficientNeural2018} as well as the kernel implementation of CMSIS-NN, as we will show in \cref{sec:experiments}.

\subsection{Implementation}

For the initial design and training of the \ac{NN}, we use TensorFlow (\texttt{v2.2.0})~\cite{TensorFlow}.
The subsequent quantization is performed with \ac{TFL}~\cite{TensorFlowLiteML}. 
The final target firmware is then compiled through the inference library \acf{TFLM}~\cite{TensorFlowLiteMicrocontrollers,davidTensorFlowLiteMicro2021} combined with the optimization library CMSIS-NN~\cite{laiCMSISNNEfficientNeural2018} as part of an Mbed OS~\cite{armltd.MbedOSn.d.} project (\texttt{v5.15.3}).
Mbed OS facilitates the dynamic deployment on a wide variety of targets featuring an ARM Cortex-M \ac{MCU} through a well-maintained compilation toolchain and the option to extend the evaluation to a full-featured embedded OS implementation.
\ac{TFLM} on the other hand is target-independent and can also be used for other CPU architectures such as RISC-V.

For the external energy measurement, we employ the open-source hardware and software project RocketLogger~\cite{SGLLLT2017}. \ac{GPIO} pins indicate the start and end of an inference as well as intermediate layers and can be traced at up to 64\,kHz.
The device logs measurements locally and can be accessed remotely; the complete test setup is automated and does not require any manual intervention.

With this toolchain, we provide the opportunity for a time-efficient evaluation of various \acp{NN} and target devices without requiring detailed technical knowledge of the implementation on \acp{MCU}. This enables engineers to focus their efforts on designing networks and systems while still being able to test and compare their results with minimal effort.
Our work, including all software and hardware components for our toolchain, is open-sourced and publicly available\footnote{\href{https://gitlab.ethz.ch/tec/public/tflm-toolchain}{gitlab.ethz.ch/tec/public/tflm-toolchain}}~\cite{heimSoftwareRepositoryToolchain2021,SGLLLT2017}.
It includes automated scripts for preparing, deploying, and analyzing \acp{NN}.
The measurement results from our evaluation are available as examples~\cite{heimMeasurementDataRepository2021}.


\section{Experiments}\label{sec:experiments}

In this section, we first investigate the non-uniform effects of optimizations on different \acp{NN}.
We then examine the relationship of the perceptible metrics (energy and latency) empirically and discuss the resulting Pareto front for our targets.
Lastly, we explore the detailed effects of layers' hyper-parameters by examining the perceptible metrics with layer granularity using artificial benchmarking \acp{NN}.

For our experiments, we select the development boards STM NUCLEO L496ZG (L4)~\cite{NUCLEOL496ZG}, STM DISCO F496NI (F4)~\cite{32F469IDISCOVERYDiscoveryKit}, and STM NUCLEO F767ZI (F7)~\cite{NUCLEOF767ZI}.
Ranging from ultra-low power (L4) over balanced (F4) to high-performance \acp{MCU} (F7), these samples represent the breadth of the ARM Cortex-M series, which we target because of its low power characteristics and ubiquity in embedded devices.
If not otherwise mentioned, all experiments are conducted on the L4 as the most resource-constrained device.
For benchmarking, the RocketLogger~\cite{SGLLLT2017} allows us to measure the energy consumption and trace target GPIOs for status information on the inference.
In our initial tests, we investigate LeNet~\cite{lecunGradientbasedLearningApplied1998} as a representation for the most constrained \acp{NN} as well as ResNet-20~\cite{heDeepResidualLearning2015} to demonstrate more demanding capabilities (see \cref{sec:appendix-arch} for the used \ac{NN} architectures).

\begin{table}[t]
  \centering
  \small
  \begin{tabulary}{\linewidth}{R C C C C C}
    \toprule
    & CMSIS-NN & TFL Acc. & TFLM Acc. & $\Delta$ Acc. & Memory Footprint [KiB]\\
    \midrule
    \textbf{U}
    & ---   & 98.79\%  & 98.79\%   & 0.00\%  & \multicolumn{1}{c}{\multirow{1}{*}{320.28}}\\
    \addlinespace[0.1cm]
    \multicolumn{1}{c}{\multirow{2}{*}{\textbf{Q}}}
    & \xmark   & 98.74\%  & 98.74\%   & 0.00\% &  \multicolumn{1}{c}{\multirow{2}{*}{85.19}}  \\
    & \cmark   & 98.74\%  & 98.76\%   & \textbf{0.02\%} &  \\
    \bottomrule
  \end{tabulary}
  \vspace{4pt}
  \caption{
  Compared to the unoptimized model~(\textbf{U}), the 8~bit quantized model~(\textbf{Q}) has a decreased accuracy by 0.05\,\% for the LeNet. However, the quantization drastically reduces the memory footprint to 26.6\,\% of the original size.
  It is noteworthy that the usage of CMSIS-NN, which can only be verified on the MCU itself, results in a slightly altered accuracy (+0.02\,\%) due to the different implementation of the underlying kernels.
  }
  \label{tab:sec4-lenet-accuracy-TFLu}
  \vspace{-0.75em}
\end{table}

\subsection{Effects of optimizations}

Leveraging our toolchain, we investigate the influence of optimizations on the perceptible metrics in an automated manner. These experiments involve all components from the initial analysis (\textbf{A}) for calculating the memory savings to the target evaluation (\textbf{F}) of the perceptible metrics.

\textbf{Quantization}
The effects of quantization can already be evaluated on the host (\textbf{B}) using the \ac{TFL} interpreter~\cite{TensorFlowLiteML}.
We achieve a memory footprint reduction of up to \SI{73}{\percent} while only losing 0.05\,\% of accuracy (\cref{tab:sec4-lenet-accuracy-TFLu}).
Consequently, when a loss in accuracy is tolerable, quantization is highly beneficial. However, this must be empirically investigated by quantizing the NN and verifying the resulting accuracy.
We also evaluate the accuracy of the model on-device by running a time-intensive verification on the MCU itself.
Unexpectedly, the quantized model using CMSIS-NN displays an increased accuracy by 0.02\,\%. 
This difference in target metrics is not a feature of CMSIS-NN, but a random effect due to specific kernel implementations and a consequence of the different underlying computations~\cite{laiCMSISNNEfficientNeural2018}. For another \ac{NN}, this might just as well result in a drop in accuracy.

\begin{figure}
  \centering
    \includegraphics[width=\linewidth]{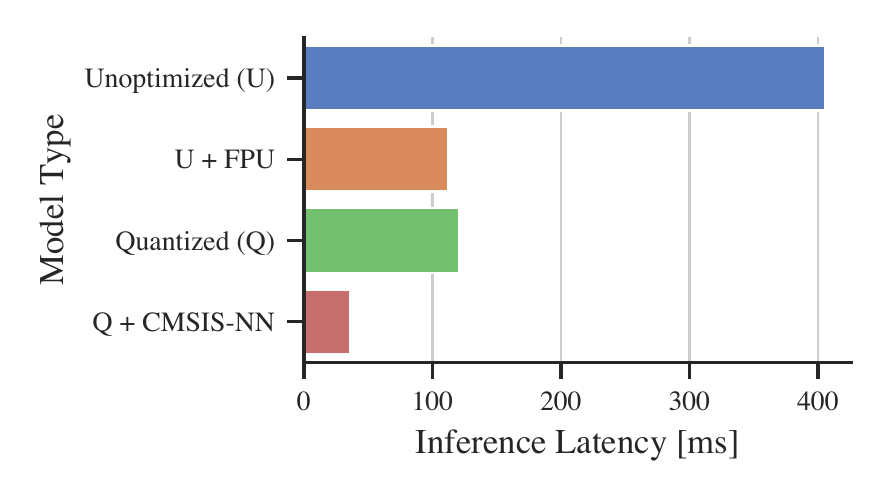}
  \caption
  {
  The combined use of 8~bit quantization (Q) and the specialized CMSIS-NN kernels result in the best performing model (Q + CMSIS-NN). However, the usage of 8~bit quantization without hardware optimizations (Q) is not faster than the floating point model (U) accelerated by the availability of an \ac{FPU} (U + FPU). 
  }
  \label{fig:sec4-latency-models}
  \vspace{-0.75em}
\end{figure}

\textbf{Inference latency}
When the \ac{FPU} is disabled, the unoptimized model (U) takes significantly longer -- using the \ac{FPU} results in a $4\times$ faster inference (U + FPU) for the LeNet.
Including the software acceleration library CMSIS-NN, we can further accelerate fixed point operations, speeding up the quantized model by an additional $4\times$ (Q + CMSIS-NN).
Consequently, even if an \ac{FPU} is available, the quantized model is significantly faster (\cref{fig:sec4-latency-models}).
Therefore, the combination of quantization and CMSIS-NN is superior in regard to memory footprint as well as latency without requiring the availability of an \ac{FPU}.
For a discussion on the effect of compiler optimizations on the latency-memory trade-off, we refer to \cref{sec:appendix-compiler}.

\textbf{Non-uniform acceleration}
When comparing the slowest and fastest model depending on the applied optimizations, we observe a consistent speedup of $13\times$ - $16\times$ across the \acp{MCU} for the LeNet (\cref{tab:sec4-latency-speedup}).
On the other hand, for the ResNet, the \ac{NN} is accelerated by $29\times$ - $35\times$. 
Consequently, it is hard to predict the effect of optimizations, as the architecture of the network and supported kernels, its complexity, and the \ac{MCU} architecture play a crucial role.

\begin{table}[t]
  \centering
  \small 
  \begin{tabulary}{\linewidth}{RcRRR}
    \toprule
    &          & \multicolumn{2}{c}{Latency [ms]} &  \\
    \cmidrule(lr){3-4}
    & MCU      & \multicolumn{1}{c}{\textbf{U}} & \multicolumn{1}{c}{\textbf{O}} &  Speedup \\
    \midrule
    \multicolumn{1}{r}{\multirow{3}[0]{*}{LeNet}}
    & L4       & {$470.5$} & {$36.4$}    & $\mathbf{12.9\times}$   \\
    & F4       & {$227.6$} & {$16.1$}    & $\mathbf{14.1\times}$   \\
    & F7       & {$126.4$} & {$8.1$}     & $\mathbf{15.7\times}$   \\
    \addlinespace[0.2cm]
    \multicolumn{1}{r}{\multirow{3}[0]{*}{ResNet}}
    & L4       & \textit{oversized}   & {$2217.0$} & \multicolumn{1}{c}{--} \\
    & F4       & {$28818.5$} & {$984.3$}  & $\mathbf{29.3\times}$   \\
    & F7       & {$15566.9$} & {$449.3$}  & $\mathbf{34.6\times}$   \\
    \bottomrule
  \end{tabulary}
  \vspace{4pt}
  \caption
  {
  Across all models and \acp{MCU}, the optimizations which lead to the fastest, optimized (\textbf{O}) and slowest, unoptimized model (\textbf{U}) are the same.
  For bigger architectures such as the ResNet, we observe a $2\times$ greater speedup compared to the LeNet.
  Furthermore, the speedup between the MCUs themselves also varies by up to 22\,\%.
  Consequently, optimizations have a substantial effect in general and benefit complex network architectures in particular.
  }
  \label{tab:sec4-latency-speedup}
\end{table}

We further find that the number of parameters is an ill-suited predictor for the computational complexity of a model. ResNet has $141\times$ more operations while only $3.4\times$ more parameters than LeNet.
To determine the predictive power of operations regarding latency, we calculate the latency ratio of LeNet and ResNet for both unoptimized (\textbf{U}) and optimized (\textbf{O}) models.
For the unoptimized models, we observe a ratio of $123\times$ - $126\times$ (\cref{tab:sec4-model-complexity}). 
However, when employing optimizations, the best performing models only demonstrate a relative difference of $55\times$ - $61\times$.
We find that optimizations result in a different scaling factor depending on the \ac{NN} architecture and employed \acp{MCU} for estimating the latency from the operations. 
Thus, the usage of proxies depends on the target and should be empirically validated.

\begin{table}[t]
  \small
  \centering
  \begin{tabulary}{\linewidth}{ccccc}
  \toprule
     & \multicolumn{2}{c}{$\frac{\mathrm{ResNet}}{\mathrm{LeNet}}$~Ratio} & \multicolumn{2}{c}{$\frac{\mathrm{ResNet~Latency}}{\mathrm{LeNet~Latency}}$} \\
  \cmidrule(lr){2-3}\cmidrule(lr){4-5}
  MCU      & \# Para.    & Ops.          & \textbf{U}         & \textbf{O}      \\ \midrule
  L4       &             &               & \textit{oversized} & ${60.9\times}$  \\
  F4       & $3.4\times$ & $141.3\times$ & ${126.6\times}$    & ${61.1\times}$  \\
  F7       &             &               & ${123.1\times}$    & ${55.8\times}$  \\
  \bottomrule
  \end{tabulary}
  \vspace{4pt}
  \caption
  {
  When optimizations (\textbf{O}) are used, the number of operations has a different scaling factor regarding its predictive quality for the inference latency.
  Furthermore, optimizations also affect the MCUs differently and lead to variations in the latency ratios.
  }
  \label{tab:sec4-model-complexity}
  \vspace{-0.75em}
\end{table}

\subsection{Energy consumption}

To evaluate the expressiveness of metrics, we investigate the correlation of the inference latency and energy consumption for the respective optimizations across the \acp{MCU}.
While we observe an ideal linear relationship ($r=0.9946$) 
between them across all optimizations, it is important to note that each \ac{MCU} has a different characteristic and hence offers an additional trade-off between latency and energy consumption. 
A look at the energy consumption of the individual layers presents a similar picture: we measure an almost linear relationship ($r=0.9995 $) between latency and energy consumption 
across our \acp{NN} and across all optimizations on a layer basis (see \cref{sec:appendix-energy}).

This observation does not match our initial assumptions, as memory access requires magnitudes more energy than computations~\cite{horowitzComputingEnergyProblem2014}.
Therefore, we assumed that layer types that lead to more memory access (e.g., dense layers) would require a disproportional amount of energy.
However, we suspect that the increased energy cost for memory access is also seen in an increased latency. As a consequence, the inference latency is a perfect proxy for the energy consumption of the investigated \acp{MCU}.
Additionally, the simple architecture of \acp{MCU} does not feature dynamic voltage scaling or power-gated sub-components of the processor, which could lead to a non-linear energy and latency relationship for more complex hardware architectures.
Accordingly, all of the presented results regarding speedups and ratios for latency also apply to energy consumption.

\begin{figure}[t]
    \centering
    \includegraphics[width=\linewidth]{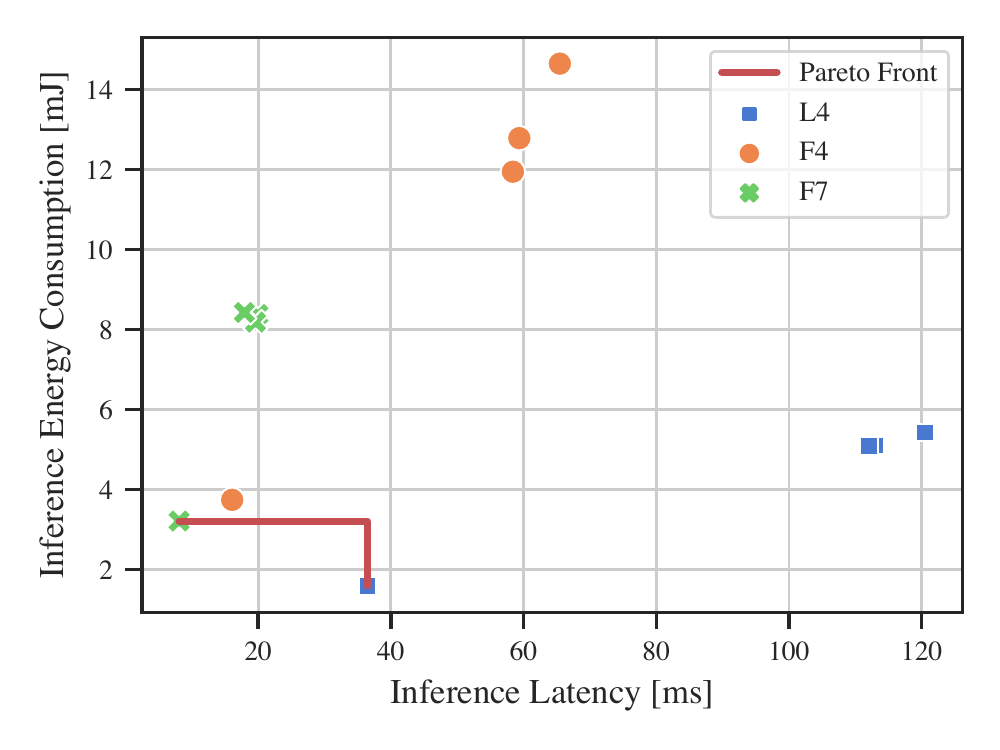}
    \caption{
    Varying the optimizations leads to different efficiencies per MCU; however, the optimizations which lead to the most efficient system are equivalent. 
    Therefore, the choice of MCU is a trade-off between the inference latency and energy consumption. While a suitable candidate depends on the actual deployment scenario, our Pareto front reveals that the F4 is not Pareto efficient.
    }
    \label{fig:sec4-pareto}
    \vspace{-0.75em}
\end{figure}

\textbf{Pareto front}
Despite the linear dependence, we find that the choice of \ac{MCU} is a trade-off between inference latency and energy consumption (\cref{fig:sec4-pareto}). Consequently, the ultimate choice of the target hardware depends on the specific operational requirements. 
We also see that not all \acp{NN} can be deployed on each \ac{MCU}, as they are limited by the available Flash memory (\textit{oversized}) or dynamic memory.
We observe in our evaluation set that one MCU (F4) is not Pareto efficient and would deliver inferior performance for any specification. 
This finding stresses the importance of empirical investigations, as the selection of a Pareto efficient \ac{MCU} results in faster inference \textit{and} less energy consumption.
As a consequence, we encourage hardware-aware testing to identify the best-fitting solution for a given scenario.

\subsection{Effects of layers' hyper-parameters}

To investigate the interplay of factors in more detail, we measure the perceptible metrics with layer granularity by sweeping over the hyper-parameters of selected layers. 
From previous work~\cite{laiCMSISNNEfficientNeural2018}, we know that CMSIS-NN is designed to leverage the underlying hardware architecture and that memory alignment plays a key role in its efficiency.
To investigate this systematically, we design benchmarking \acp{NN} which primarily consist of the layer type of interest. The hyper-parameters are iteratively incremented and deployed using our toolchain to observe their influence on the perceptible metrics.
This enables us to extract guidelines for the efficient design of \acp{NN} to holistically increase the synergy between \ac{NN} design, software libraries and hardware.

\begin{figure}[t]
    \centering
    \includegraphics[width=\linewidth]{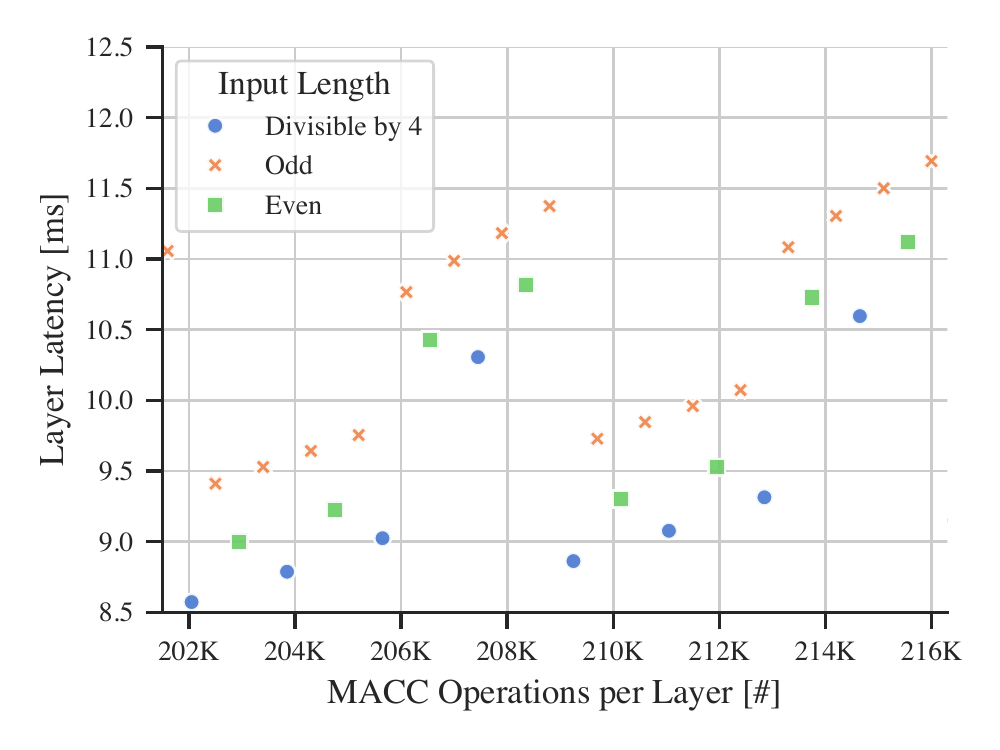}
    \caption{
    For dense layers, we observe a non-linear relationship between the number of operations per layer and its latency, with overlapping periodic patterns due to loop executions and memory accesses.
    Depending on whether the input length is \textit{odd}, \textit{even}, or \textit{divisible by 4}, more operations do not necessarily increase latency.
    }
    \label{fig:sec4-dense-flops}
    \vspace{-0.75em}
\end{figure}

\textbf{Dense layers}
When investigating the effect of the number of units for dense layers, we find that it is not their quantity that affects the perceptible metrics non-linearly but the number of \textit{input connections} per dense unit. These inputs are consecutively accessed in memory, which is favorable for efficiency. 
Consequently, we observe a non-linear relationship between them and the inference latency (\cref{fig:sec4-dense-flops}).
We find that this effect is sufficiently pronounced that we can achieve faster inference despite more operations by increasing the number of inputs to be even or a multiple of 4. 
However, we also observe a periodicity of 16 with two distinct clusters of sizes 9 and 7, whereby the first cluster is consistently more efficient.
Therefore, we can increase the number of operations while decreasing latency, which simultaneously boosts accuracy and reduces perceptible metrics.
This highly non-linear behavior reveals the shortcomings of operations as a proxy metric for \ac{NN} complexity.

\textbf{2D convolutional layers}
For convolutional layers, we similarly find that the input is decisive.
While dense layers only consist of a single dimension as they are flattened, convolutional layers have multiple dimensions 
which depend on the preceding layer.
We investigate the effect of the kernel size, the stride, and the number of filters of a preceding layer, as those hyper-parameters determine the dimensions of the output, and consequently the next input.

\begin{figure}[t]
    \centering
    \includegraphics[width=\linewidth]{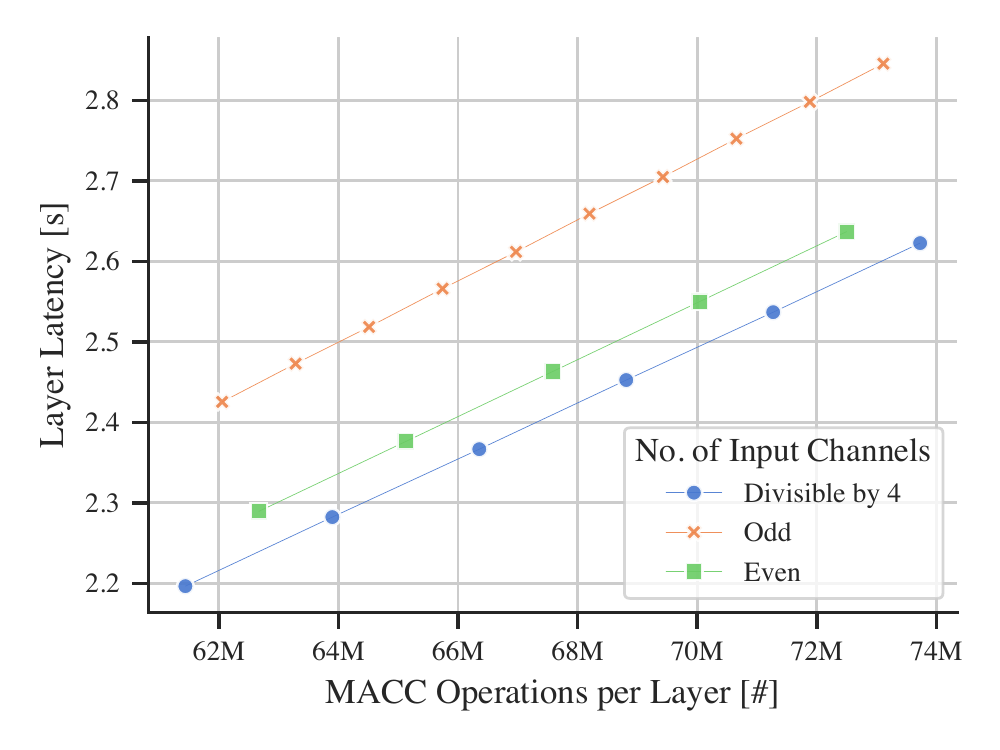}
    \caption{
    For convolutional layers, the input channels 
    are the determining hyper-parameter.
    If the number of input channels is \textit{divisible by 4}, 8~bit memory access of the 32~bit architecture is aligned and we observe a decreased latency and energy consumption despite an increase in the number of operations.
    }
    \label{fig:sec4-conv-flops}
    \vspace{-0.75em}
\end{figure}

We find that the length of the input channels is the key hyper-parameter, which is determined by the preceding layer's number of filters.
To verify this, we create a network with a static convolutional layer which we benchmark and a preceding convolutional layer where we iteratively increase the number of filters.
Up to two simultaneous operations can be executed by leveraging SIMD instructions if the number is \textit{even}, which we verify through an inspection of the source code.
While SIMD offers the potential for up to four parallel computations, the number of MCU registers limits its exploitation in this case.
Despite an increasing number of operations, we observe a reduction in our perceptible metrics if the number of input channels is even or divisible by 4, enabling us to accelerate computation by up to 7.7\,\% by adding a filter to the previous layer (\cref{fig:sec4-conv-flops}).

\textbf{Latency per operation}
The latency per operation $\delta$ allows us to compare layers and their hyper-parameters directly based on perceptible metrics. 
We see that aligned memory access and the exploitation of SIMD instructions result in significant gains $G$ (\cref{tab:sec4-utilization}). 
The usage of CMSIS-NN ($\delta_{Optimized}$) decreases the latency across all layer types compared to the hardware-agnostic implementation 
($\delta_{NC}$).
However, whether these techniques can be leveraged depends on the layer type. For example, depth-wise convolution cannot make use of aligned memory access due to 
the required, predetermined structure of the dimensions in memory.
As a result, the latency of operations depends on the layer type, its hyper-parameters, and the optimizations and cannot be abstracted for proxy metrics.

\begin{table}
  \footnotesize
  \begin{tabulary}{\linewidth}{RR|RRRR}
  \toprule
                & \multicolumn{1}{c}{$\delta_{NC}$} & \multicolumn{4}{c}{$\delta_{Optimized}$} \\
    \cmidrule(lr){3-6}
    Layer Type  &  \multicolumn{1}{c}{[\SI{}{\nano\second \per Op}]} & \multicolumn{1}{c}{$G_{all}$} & \multicolumn{1}{c}{$G_{odd}$} & \multicolumn{1}{c}{$G_{even}$} & \multicolumn{1}{c}{$G_{\%4}$} \\
    \midrule
    Dense        &  $\SI{145.2}{}$ & $3.00\times$  &  $2.89\times$  & $3.13\times$ & $3.24\times$  \\
    2D Conv.     &  $\SI{250.6}{}$ & $6.70\times$  &  $6.42\times$  & $6.96\times$ & $7.03\times$  \\
    1x1 2D Conv. &  $\SI{148.5}{}$ & $6.05\times$  &  $5.72\times$  & $6.33\times$ & $6.60\times$  \\
    DW 2D Conv.  &  $\SI{925.1}{}$ & $2.23\times$  &  $2.17\times$  & $2.29\times$ & $2.32\times$  \\ 
    \bottomrule
  \end{tabulary}
  \vspace{4pt}
  \caption{
  The computational performance differs depending on the hyper-parameters. 
  Most notably is the increased gain $G$ of the optimized ($\delta_{Optimized}$) compared to the unoptimized ($\delta_{NC}$) models.
  However, for certain layers like depth-wise (DW) 2D convolution, an efficient exploitation of the hardware is less effective.
  \textit{Odd}~($G_{odd}$), \textit{even}~($G_{even}$), and \textit{divisible by 4}~($G_{\%4}$) corresponds to the input length or number of input channels, respectively.
  }
  \label{tab:sec4-utilization}
  \vspace{-0.75em}
\end{table}

\subsection{Summary}

Instead of optimizing layers separately, the interplay of layers 
reveals to be crucial.
Consequently, the influence of optimizations and hyper-parameters is interdependent and a result of layer types and dimensions.
However, while some layers can greatly benefit from these techniques 
(e.g., dense and convolution), others (e.g., depth-wise convolution) are limited in performance due to the nature of their operation 
and their efficiency gain is smaller.
The effective costs of layers should therefore already be included \textit{during} the design process when choosing layer types and dimensions.


\section{Discussion}\label{sec:discussion}

Even when building on top of existing frameworks, on-device inference and testing still require significant engineering and a deep understanding of embedded systems. We try to address this shortcoming to lower the entry burden for \ac{ML} domain expert and facilitate empirical studies. 

As presented in \cref{sec:experiments}, the development of a heuristic with metrics known prior to deployment is challenging -- especially when optimizations and hardware features are employed. The introduction of further optimizations such as the exploitation of sparsity~\cite{fedorovSpArSeSparseArchitecture} is likely to make matters even more complex.
Operations can often be used as a rough general-purpose proxy for comparing \acp{NN}. 
However, especially with only subtle differences in the number of operations, it is difficult to obtain reliable efficiency estimations without implementation-aware experiments as perceptible metrics can vary significantly. Additionally, sometimes more operations can lead to decreased energy consumption and latency.
Therefore, we find the verification and evaluation via deployment on hardware indispensable and present a methodology to investigate this empirically.

\subsection{Design guidelines}
Based on our insights, we propose a more nuanced understanding of operations and therefore recommend:
\begin{itemize}
	\item The use of quantization with a hardware-supported bit-width (e.g., 8~bit) to leverage the underlying architecture, reduce memory size and accelerate inference. 
	\item An understanding of the kernel implementation to choose efficient layer types and dimensions (e.g. for \ac{TFLM} on Cortex-M devices with CMSIS-NN, a number of input channels which is divisible by 4).
	\item The verification of proxy metrics by empirically measuring them and demonstrating their merit for a specific NN on the application-relevant target. Perceptible metrics are what really matters -- not proxies.
\end{itemize}

\subsection{Future work}

Our findings could be directly applied to the manual design of efficient \ac{NN} architectures.
Furthermore, our toolchain can also be incorporated in NAS to explore an optimized search space dimension more efficiently (e.g. by avoiding known inferior parametrizations).
The search space itself can already take the hardware architecture and quantization bit-width into account by modifying the hyper-parameters accordingly and hence be reduced in size.

Additionally, our work permits researchers to evaluate candidate networks directly on perceptible metrics. 
We have found that the latency and energy consumption are independent of the actual value of the weights, as computation time and memory access patterns remain identical. 
Therefore, untrained candidate networks 
can already be efficiently evaluated before the computationally expensive training. Candidates which do not fit strict energy or latency criteria can be immediately discarded.


\section{Conclusions}\label{sec:conclusions}

This paper demonstrates that state-of-the-art \acp{NN} can be efficiently deployed on resource-constrained devices.
We show that the usage of optimizations, both generic and hardware-specific, can significantly decrease the memory footprint of \acp{NN} and accelerate their inference latency.
We further demonstrate that empirical investigations are indispensable due to the variability and the interdependence of \ac{NN} layers in both software and hardware.
We have found that empirical implementation-aware verification
is the only reliable method to obtain key performance metrics. 
Therefore, our developed toolchain is a cost-effective method for researchers and engineers to optimize, investigate, deploy, and benchmark \acp{NN} on ARM Cortex-M devices.
In addition, we provide insights and guidelines for the design of efficient \acp{NN}.
With this methodology which can be directly incorporated as a NAS component, we expect to see more efficient TinyML applications designed in symbiosis with their targeted application scenario through an implementation-aware development process. 



\vspace{-0.3em}

\section*{Acknowledgments}

This research was supported by the Swiss National Science Foundation under NCCR Automation,
as well as a research grant of the IDEA League and the IFI program of the German Academic Exchange Service (DAAD).

\begin{acronym}
    \acro{IoT}{Internet of Things}
    \acro{ML}{Machine Learning}
    \acro{AI}{Artificial Intelligence}
    \acro{ASP}{average selling price}
    \acro{FLOP}{floating point operation}
    \acro{MACC}{multiply accumulate}
    \acro{SQNR}{signal-to-quantization-noise ratio}
    \acro{NN}{neural network}
    \acro{ANN}{artificial neural network}
    \acro{DNN}{deep neural network}
    \acro{CNN}{convolutional neural network}
    \acro{DCN}{deep convolution network}
    \acro{FFT}{fast Fourier transform}
    \acro{ReLU}{rectified linear unit}
    \acro{GPU}{graphics processing unit}
    \acro{MCU}{microcontroller unit}
    \acro{DSP}{digital signal processor}
    \acro{TPU}{tensor processing unit}
    \acro{NAS}{neural architecture search}
    \acro{SIMD}{single instruction multiple data}
    \acro{LSTM}{long short-term memory}
    \acro{SoC}{system on chip}
    \acro{RAM}{random-access memory}
    \acro{COTS}{commercial off-the-shelf}
    \acro{CMSIS}{Cortex Microcontroller Software Interface Standard}
    \acro{CMSIS-NN}{\ac{CMSIS}-Neural Networks}
    \acro{TF}{Tensor\-Flow}
    \acro{TFL}{Tensor\-Flow Lite}
    \acro{TFLM}{Tensor\-Flow Lite Micro}
    \acro{FPU}{floating point unit}
    \acro{OS}{operating system}
    \acro{GPIO}{general purpose input/output}
    \acro{LSB}{least significant bit}
    \acro{MSB}{most significant bit}
    \acro{IC}{integrated circuit}
    \acro{RTOS}{real-time operating system}
    \acro{OTA}{over-the-air}
    \acro{DFU}{device firmware upgrade}
    \acro{LPWAN}{low-power wide-area network}
    \acro{ASIC}{application-specific integrated circuit}
    \acro{ASIP}{application-specific instruction set processor}
\end{acronym}

{\small
\bibliographystyle{format/ieee_fullname}
\balance
\bibliography{bibliography}
}

\clearpage
\nobalance

\appendix

\section{Calculating FLOPs and MACC operations}\label{sec:appendix-calcops}

\subsection{Dense layer}

The total number of required arithmetic operations can be expressed in \acp{FLOP}:
\begin{equation}
\begin{split}
	\mathrm{FLOPs}_{DL} = N_{out} N_{in} + N_{out} (N_{in}-1) + N_{out}\\
	 = 2~N_{out}N_{in} = 2~(\mathrm{Paras}_{DL} - N_{out}).
\end{split}
\end{equation}
$N_{out}$ additions are required for the bias term, along with the $N_{out} \cdot N_{in}$ multiplications. Lastly $N_{out} \cdot (N_{in} - 1)$ additions are used for the accumulation results within a neuron.

The usage of \ac{MACC} operations\footnote{A \ac{MACC} computes the product of two numbers and accumulates the result: $ a \leftarrow a + (b \cdot c)$.} allows one to already accumulate the multiplications for each individual neuron -- therefore it results in $N_{out} \cdot (N_{in} - 1)$ less operations:
\begin{equation}
	\mathrm{MACCs}_{DL} = N_{out}~(N_{in}+1) = \mathrm{Paras}_{DL}.
\end{equation}

\subsection{Convolutional layer}

For the following equations, we assume that the convolution is implemented as a sliding window.
Compared to the number of parameters, the computational complexity in \acp{FLOP} or \acp{MACC} strongly depends on the size of the input:
\begin{equation}
\begin{split}
	\mathrm{FLOPs}_{CL} \approx
	\underbrace{\left(\frac{I_x - K_x + 2P_x}{S_x} + 1\right)}_\text{{output size}~x} \times \\
	\underbrace{\left(\frac{I_y - K_y + 2P_y}{S_y} + 1\right)}_{\text{output size}~y} \times \\
	~ C_{in} 
	\underbrace{\left( 2~K_x K_y  + 1 \right)}_{\substack{\text{FLOPs} \\ \text{per output element}}}
	C_{out},
\end{split}
\end{equation}
where $I_d$ is the length of the input, $P_d$ the padding length, and $S_d$ the stride in their respective dimension $d$.
In \ac{MACC} operations, the convolution can be accelerated as the multiplication and accumulation for applying the filter can be done in a single operation:
\begin{equation}
\begin{split}
	\mathrm{MACCs}_{CL} \approx
	\underbrace{\left(\frac{I_x - K_x + 2P_x}{S_x} + 1\right)}_\text{{output size}~x} \times \\
	\underbrace{\left(\frac{I_y - K_y + 2P_y}{S_y} + 1\right)}_{\text{output size}~y} \times \\
	~ C_{in} 
	\underbrace{\left(K_x K_y + 1 \right)}_{\substack{\text{MACCs} \\ \text{per output element}}}
	C_{out}.
\end{split}
\end{equation}
Assuming no padding and a typical stride of $1$, the formula can be simplified to:
\begin{equation}
\begin{split}
	\mathrm{MACCs}_{CL} \approx (I_x - K_x + 1)~(I_y - K_y + 1) \\
	C_{in} (K_x K_y + 1)~C_{out}.
\end{split}
\end{equation}

\subsection{Depth-wise convolutional layer}

For the depth-wise convolution, the number of FLOPs can be calculated as follows: 
\begin{equation}
\begin{split}
	\mathrm{FLOPs}_{DWCL} \approx
	\left(\frac{I_x - K_x + 2P_x}{S_x} + 1\right) \times \\
	\left(\frac{I_y - K_y + 2P_y}{S_y} + 1\right) \times \\
	C_{in} (2 K_x K_y + 1).
\end{split}
\end{equation}
%
%
Similarly, the number of MACC operations can be derived as:
\begin{equation}
\begin{split}
	\mathrm{MACCs}_{DWCL} \approx
	\left(\frac{I_x - K_x + 2P_x}{S_x} + 1\right) \times \\
	\left(\frac{I_y - K_y + 2P_y}{S_y} + 1\right) \times\\
	C_{in} ( K_x K_y + 1).
\end{split}
\end{equation}
Assuming a stride $S$ of $1$ and no padding $P$, this can be reduced to:
\begin{equation}
\begin{split}
	\mathrm{MACCs}_{DWCL} \approx (I_x - K_x + 1)~(I_y - K_y + 1) \\
	C_{in} (K_x K_y + 1).
\end{split}
\end{equation}

\section{Additional experiments}\label{sec:appendix}


\subsection{Compiler optimizations}\label{sec:appendix-compiler}

We find that the inference latency can be decreased at the cost of increased code size by exploiting compiler optimizations.
We observe a reduction of 115\,KiB using the compiler flag \texttt{-Os}, ordering the compiler to optimize for size, which is a 31\,\% reduction for the LeNet deployment, but only a 19\,\% reduction for the ResNet.
This is due to the fact that optimizing for size offers diminishing benefits if the majority of the binary file consists of the \ac{NN}, as code optimization only applies to the libraries but not to the network parameters themselves.
Therefore, we recommend using \texttt{-Ofast} when the \ac{NN} makes up the majority of the binary size, as this improves the inference speed by a significant factor ($13\,\%$ - $29\,\%$ for the quantized CMSIS-NN model) without substantially affecting the binary size.

\newpage
\subsection{Energy and latency relationship}\label{sec:appendix-energy}

\begin{figure}[!ht]
   \centering
   \includegraphics[width=\linewidth]{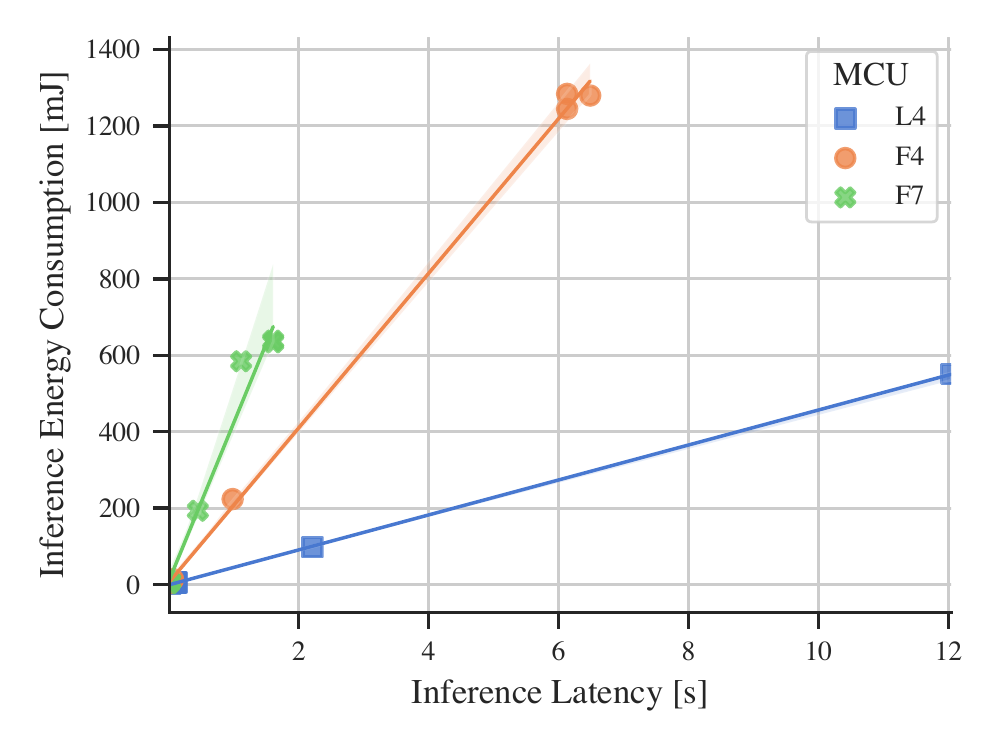}
   \caption{
   We measure the inference latency and energy consumption for complete \ac{NN} executions with different combination of optimizations for the LeNet and ResNet.
   They display a perfect linear relationship ($r = 0.9946$) with varying slopes for the \acp{MCU}, showing different latency and energy consumption characteristics.
   }
   \label{fig:appendix-energy-regression-models}
\end{figure}

\begin{figure}[!ht]
   \centering
   \includegraphics[width=\linewidth]{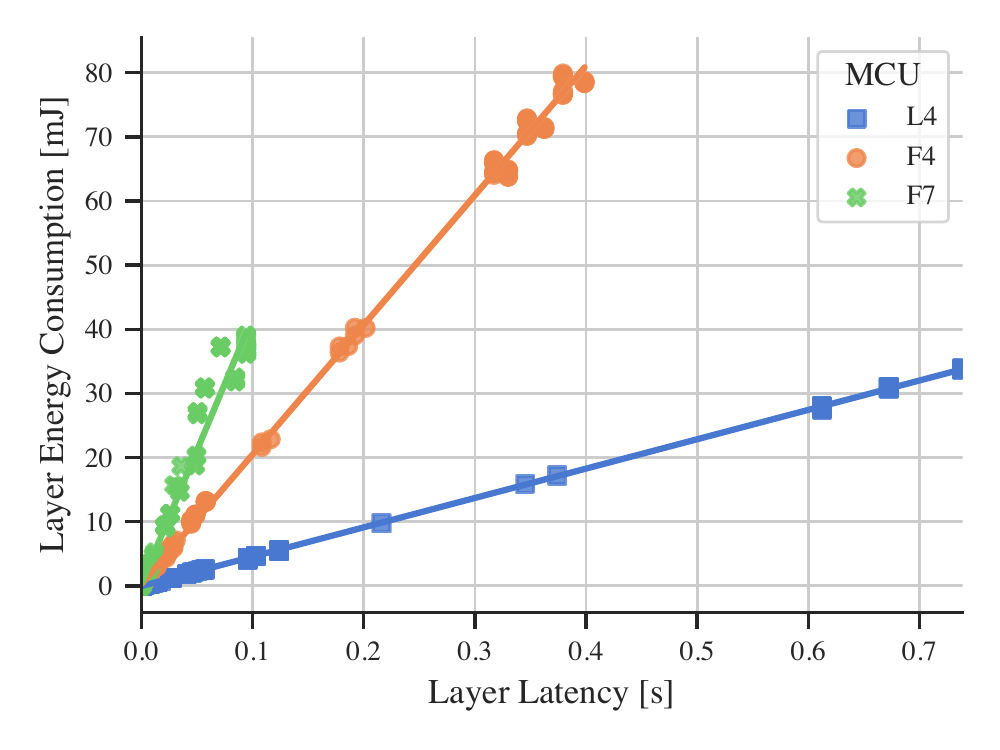}
   \caption{
   We observe a linear relationship ($r = 0.9995$) for the latency and energy consumption of the individual layers of the LeNet and ResNet across all \acp{MCU}, independent of the layer type.
   }
   \label{fig:appendix-energy-regression-layers}
\end{figure}

\newpage
\subsection{Effects of layers' hyper-parameters}\label{sec:appendix-layer}

\begin{figure}[!ht]
   \centering
   \includegraphics[width=\linewidth]{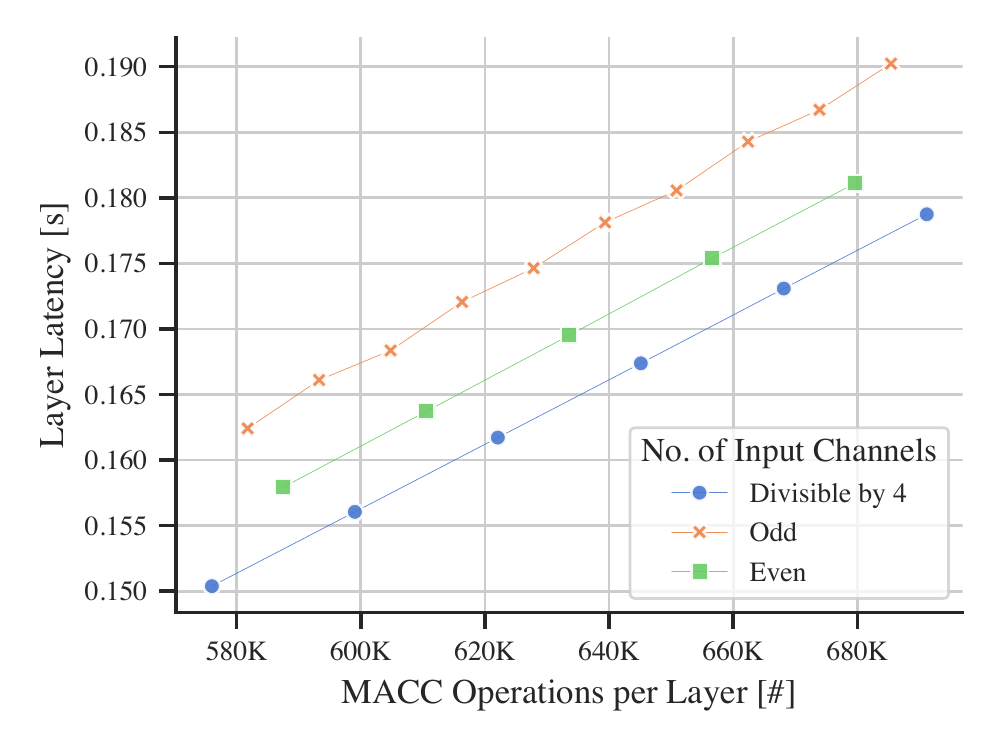}
   \caption{
   Depth-wise convolution has a similar characteristic as the convolutional layer.
   However, the overall latency per operation exceeds all other layers and does not significantly decrease if the hardware is utilized (\cref{tab:sec4-utilization}). Consequently, the latency cost per operation is substantially higher.
   }
   \label{fig:appendix-DW-conv}
\end{figure}

\begin{figure}[!ht]
   \centering
   \includegraphics[width=0.8\linewidth]{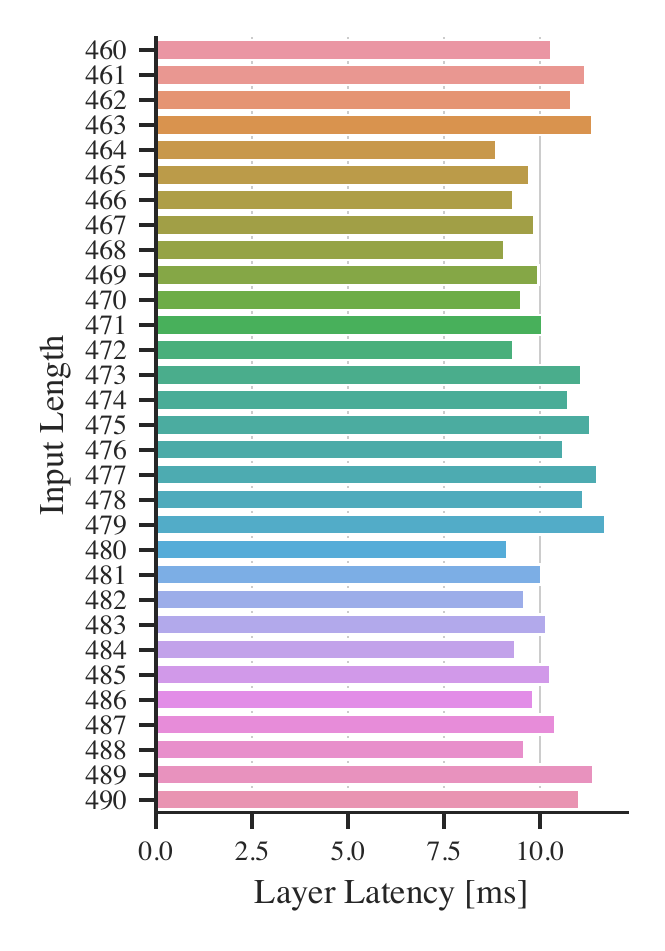}
   \caption{Raising the number of inputs does not lead to a linear increase in latency for a static dense layer, but displays a pattern of overlapping periods depending on a combination of SIMD instruction usage and multiple nested loops.}
   \label{fig:appendix-dense-latency}
\end{figure}

\newpage
\section{Neural network architectures}\label{sec:appendix-arch}

\begin{figure}[!ht]
   \centering
   \includegraphics[width=1.0\linewidth]{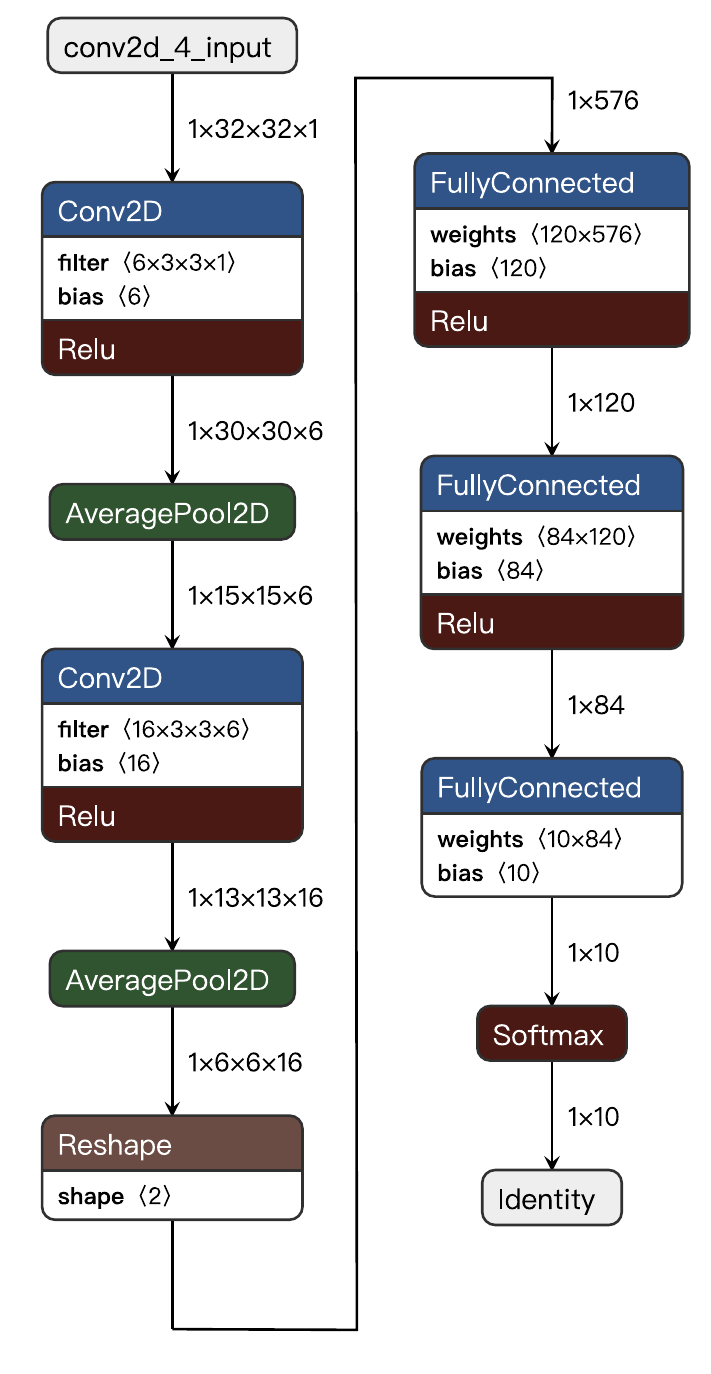}
   \caption{LeNet architecture.}
   \label{fig:appendix-NN-LeNet}
\end{figure}

\begin{figure}[!ht]
   \centering
   \includegraphics[width=.85\linewidth]{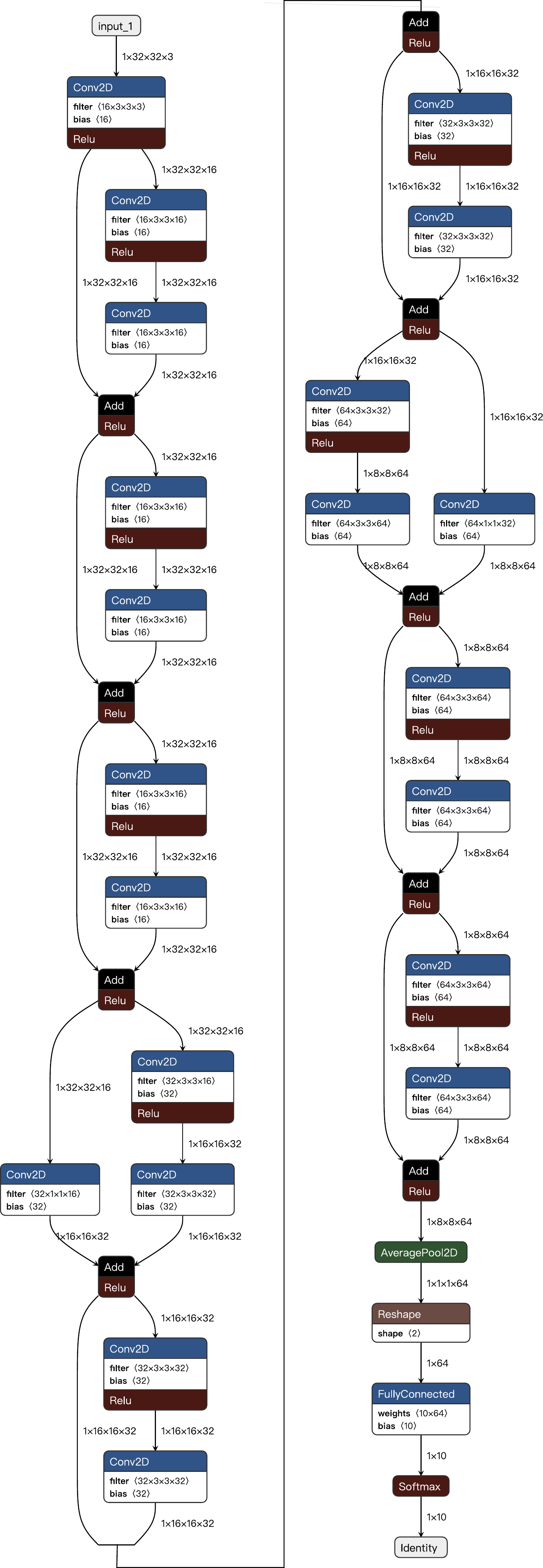}
   \caption{ResNet-20 architecture.}
   \label{fig:appendix-NN-ResNet}
\end{figure}

\begin{figure}[!ht]
   \centering
   \includegraphics[width=.36\linewidth]{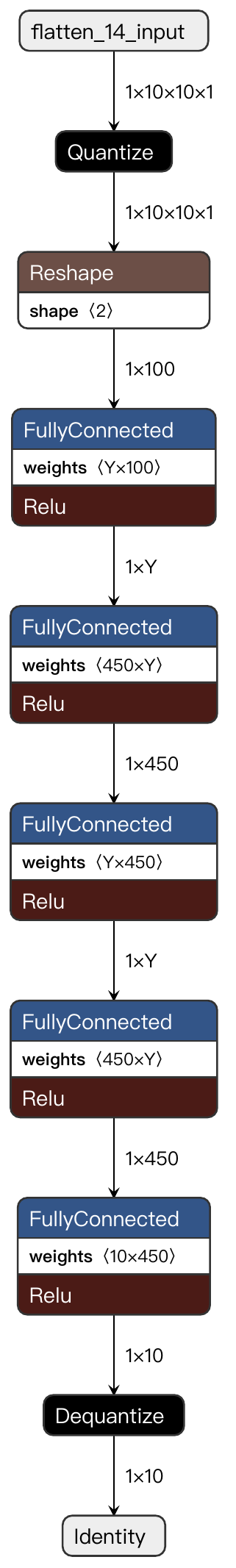}
   \caption{NN architecture for the dense layer benchmarking network. $Y$ denotes the iteratively incremented variable.}
   \label{fig:appendix-NN-dense}
\end{figure}

\begin{figure}[!ht]
   \centering
   \includegraphics[width=.5\linewidth]{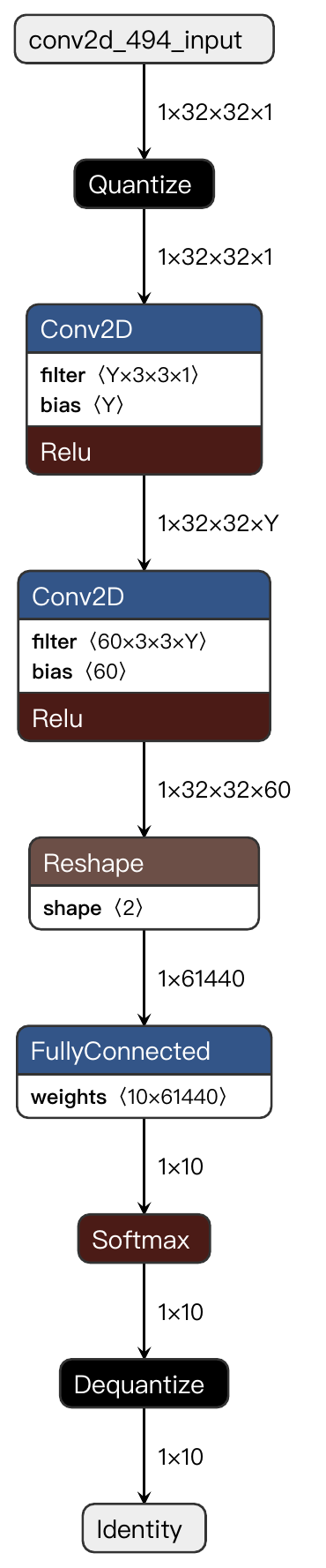}
   \caption{NN architecture for the convolutional layer benchmarking network. $Y$ denotes the iteratively incremented variable.}
   \label{fig:appendix-NN-conv}
\end{figure}

\begin{figure}[!ht]
   \centering
   \includegraphics[width=.5\linewidth]{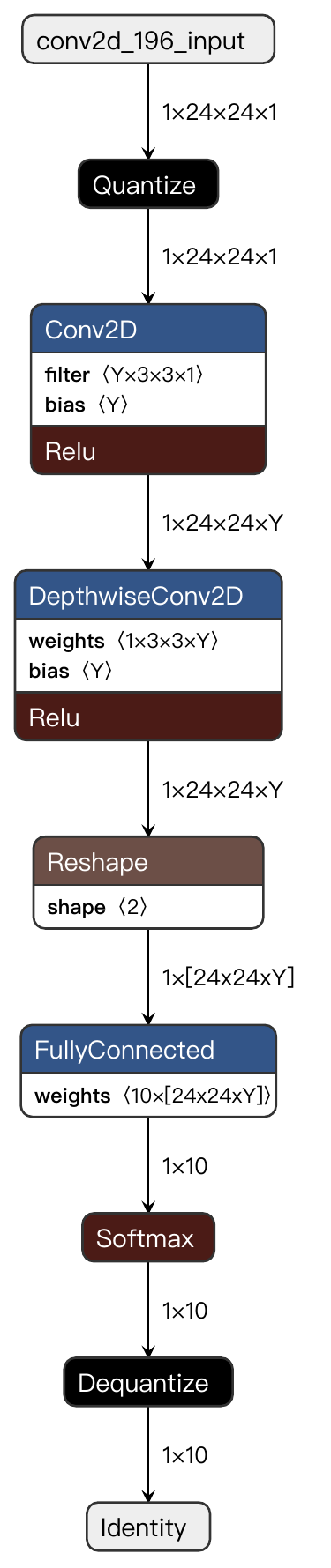}
   \caption{NN architecture for the depth-wise convolutional layer benchmarking network. $Y$ denotes the iteratively incremented variable.}
   \label{fig:appendix-NN-conv-dw}
\end{figure}

\end{document}